# Dance Teaching by a Robot: Combining Cognitive and Physical Human–Robot Interaction for Supporting the Skill Learning Process

Diego Felipe Paez Granados[1], Breno A. Yamamoto[2], Hiroko Kamide[3], Jun Kinugawa[1], and Kazuhiro Kosuge[1]

*Abstract*—This letter presents a physical human–robot interaction scenario in which a robot guides and performs the role of a teacher within a defined dance training framework. A combined cognitive and physical feedback of performance is proposed for assisting the skill learning process. Direct contact cooperation has been designed through an adaptive impedance–based controller that adjusts according to the partner's performance in the task. In measuring performance, a scoring system has been designed using the concept of *progressive teaching* (PT). The system adjusts the difficulty based on the user's number of practices and performance history. Using the proposed method and a baseline constant controller, comparative experiments have shown that the PT presents better performance in the initial stage of skill learning. An analysis of the subjects' perception of *comfort, peace of mind*, and *robot performance* have shown significant difference at the $p < .01$ level, favoring the PT algorithm.

*Index Terms*—Physical Human–Robot Interaction, Force Control, Education Robotics, Human Factors and Human-in-the-loop, Human Performance Augmentation.

## I. INTRODUCTION

HUMAN–robot interaction (HRI) under the context of teaching has been studied in recent years for the possibility of dedicating the same level of attention to each student during the learning process. Robot companionship has shown to benefit the students in multiple aspects, e.g., increasing the students interest and attention levels in classrooms [1] and enhancing the learning process rates in language lessons [2]. These approaches combine continuous following of the pupil progress state within the established task, teaching with novel feedback methods that motivate students, and incentivizing further learning. In these approaches, the use of a robot provides benefits such as real-time tracking of the learning process and continuous assessment and feedback, which are important aspects for significant learning [3].

In parallel to the interest in enhanced teaching strategies through HRI, recent studies of physical human–robot interaction (pHRI) intend to achieve better human assistance during direct contact situations. Multiple studies have shown that haptic assistance during physical cooperative work with a robot benefit from balancing the leading/following role of the robot [4]–[7]. These studies introduce novel adaptations that rely mostly on the concept of impedance control [8]. Adding continuous role change of the robot [4]–[6], or through a proportional-type control based on performance changes [7]. All these reports agree that cooperation cannot be achieved through unilateral input from humans, thus, robots should be companions rather than slaves [9]. Although there is not yet a general agreement on the policies for adapting the robot behavior, adapting the control parameters has shown to enhance success on the proposed tasks.

In this work, we set a dance framework with a robot teaching humans to perform a social dance. The framework takes advantage of a predefined leader/follower role assignment (which facilitates couple synchronization [10], [11]). In contrast to previous works, the human student would take the follower role (from the high-level cognitive interaction) permitting the robot to guide motions toward teaching dance figures (combination of a sequence of steps). Although the robot relies on this assumption of human compliance during the interaction for guiding (the partner should react to robot guided motions), it should accommodate and respond in accordance with the partner's ability and performance.

We propose to use an adaptive impedance control based on cumulative performance and progressive task difficulty for enhancing the learning process. The robot should perform a continuous adaptation of the dynamics of interactions for enhancing the human internal skill modeling. This proposal follows the findings of higher retention rates of learning skills through amplified dynamics reported for rehabilitation scenarios [12], as well as findings of assistance policies introduced for haptic guidance in virtual environments [13]. Our approach focuses on training of a more complicated skill (dancing); therefore, feedback to students of the knowledge of performance (KP), and knowledge of results (KS) plays a fundamental role in the learning process [14]. Consequently, the robot supports the *cognitive learning stage* during training through dynamic feedback of performance combined with haptic guidance.

This study has two main contributions. First, it provides a pHRI example where the human student assumes the follower role resulting in a perceived comfortable and non-stressful interaction for the student as shown by subjective evaluations. Second, the study demonstrates a long-term adaptive controller based on the progressive teaching (PT) methodology, which is evaluated through a cumulative performance scoring (CPS) system that provides dynamic feedback to the student.

This paper is structured as follows. Section II proposes the control architecture of the robot for close contact interactions. Section III presents the teaching methodology and assessment system. Section IV explains the experimental setup together with quantitative and qualitative evaluations. Section V addresses the results of training with subjects and the interaction perceptions. Section VI discusses the present and future works.

## II. MOTION CONTROL UNDER CLOSED CONTACT pHRI

### A. Dance Framework for Close Contact Interaction

Previous works on pHRI within dancing include the Partner Ballroom Dance Robot project [11], [15], which aimed to understand and predict human motions toward a natural pHRI. Controlling the interaction through a coupled double inverted pendulum model of the human–robot system, the project achieved a satisfactory interaction where the robot could predict, and react appropriately to human stimulus. In contrast, we propose to develop a robot that guides the partner's motion within a training dance scenario. Unlike the task of reacting to human stimulus reported in previous works. It requires the robot to convey and guide the motion direction (dance figure) through force interaction while complying with errors in the trajectory. In particular, for teaching novice dance partners, guidance of the motion requires a high level of force to reduce the couple error. Nevertheless, such guidance should be limited to ensure the safety of the partner (which we define in this context as the perceived level of disturbance to the mental state of the human partner), so that the partner can concentrate in learning.

With this objective, we have developed a new robot, a Dance Teaching Robot (DTR), whose physical configuration is presented in Fig. 1. It is a mobile humanoid $1.8m$ high, composed by a mobile base with three omni-directional wheels and four actuated joints on the upper body: $q_1$, linear motion joint for rise and fall; $q_2$ a torso rotation joint; $q_3$, a joint for rotation of the upper torso; and $q_4$, joint for rotation of the arms around the neck. These joints combine to present a human-like motion for the dance and allow the robot to exert forces to a human partner in order to guide motions. A six-axis force/torque sensor placed below joint $q_2$ senses the resulting interaction forces and torques, which are used to guide the whole-body motions. A design based on a human-like approach to the physical communication for ballroom dance, where the arms should be immobile forming the *arms frame*, so that the main means of communication in this framework is the partner's COM motion perceived by the force interaction at the hip contact [16]. Finally, two LRFs are placed at the mobile base for legs motion tracking.

The robot–human contact points are shown in Fig. 1; designed for contact with adults with height range $1.5 - 1.9m$. So that the robot embraces the human with the right arm and contacts the human by the left hand and at the hip level, creating the dance *closed position* [17]. Such close contact creates the need to provide comfort (here defined as the perceived sensation of freedom from constraints during motions and absence of distress), and perceived safety to the partner. Thus, the system should create an interaction that permits the partner to concentrate on the learning of the motion itself rather than the robot.

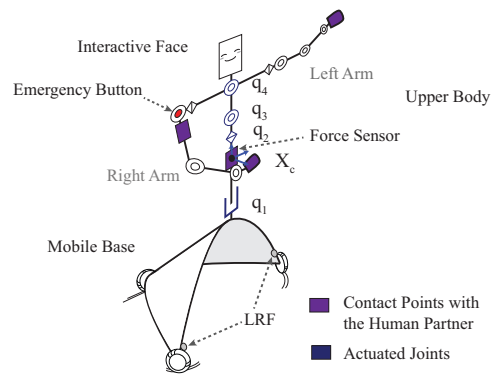

Fig. 1. Dance Teaching Robot joints and contact points.

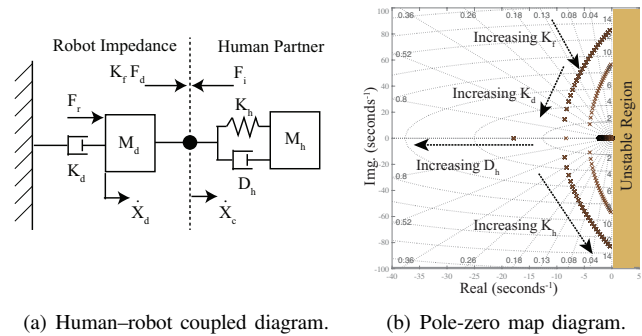

(a) Human–robot coupled diagram. (b) Pole-zero map diagram.

Fig. 2. Impedance model and its stability response for variations in human stiffness $K_h$ and damping $D_h$ within a range of robot $K_d$ and $K_f$.

### B. Low-level Interaction Control

To fulfill the requirements described in the previous section for the close interaction, a controller was designed based on an impedance model [8]. However, in dancing with a mobile robot, stiffness in the model could endanger the balance of the couple, as normal small deviations occur at every step, which might induce instability in the motion. Therefore, we defined a model without stiffness (Fig. 2(a)), as introduced in previous studies [18]–[20]. In this case, the model responds to a desired direction of motion (velocity), permitting a steady-state error in position.

We limited the interaction force $F_i$ by establishing a desired interaction force $F_d$ at the contact point. This scheme permits us to establish a boundary to the robot guidance, as it could produce large interaction forces uncomfortable for the human partner.

The human counterpart is considered as an inertial load $M_h$ attached to the robot through a spring-damper ($K_h$, $D_h$) system. As shown in [19], humans can modify their body stiffness, and compliance at will, executing tasks in a wide range of abilities. Accordingly, multiple studies have analyzed the stability of this model for HRI [18], [20]. We present in Fig. 2(b) a root locus analysis through a pole map of the closed loop system for this model based on the previous work in [20]. For robot parameters in the $X$ axis described in section IV-A and variations of human parameter $0 \leq Kh \leq 500 KN/m$, $0 \leq Dh \leq 1.2 KNs/m$, with constant inertia $M_h = 70 Kg$. This result shows that very high human stiffness would drive the system to instability, for known bandwidth limits of the robot model.

Nevertheless, in the dance training scenario, we rely on human passivity for the interaction [21], i.e., in this task, the human partner has a predefined role to comply. Therefore, human stiffness $K_h$, the possible cause of an unstable response, would operate in low levels. Under this condition, control parameters of force gain $K_f$ and damping gain $K_d$ are set within bandwidth limits of the robot internal dynamic model. The general control law for the robot can be written as

$$M_d(\ddot{X}_c - \ddot{X}_d) + K_d(\dot{X}_c - \dot{X}_d) = K_f F_d - F_i, \quad (1)$$

where $F_d$ represents the desired interaction force, $F_i$ the measured interaction force with the human partner, both in $\mathbb{R}^6$. $\dot{X}_d$ describes the desired velocity, and $\dot{X}_c$ the real velocity of the coupled motion. With an internal dynamic model of the robot,

$$F_r = M_r \ddot{X}_c + \eta_r + F_i, \quad (2)$$

where $F_r$ denotes the actuators forces which are related to output torques as $\tau_r = J_r^T F_r$, for a Jacobian $J_r$ relating the actuators space $q_i \in \mathbb{R}^{n_{dof}}$ to the contact space $X_c \in \mathbb{R}^6$, in this case, $n_{dof} = 7$. $M_r$ stands for robot inertia and $\eta_r$ represents the non-linear dynamics (Coriolis, and centrifugal components). Combining (1) and (2), and setting $M_d$ to the actual robot inertia $M_r$, the force control law can be written as

$$F_r = M_r \left( \ddot{X}_d + M_r^{-1}(K_d(\dot{X}_d - \dot{X}_c) + K_f F_d - F_i) \right) + \eta_r + F_i, \quad (3)$$

This control law permits the partner to stop the robot guidance at the desired interaction force limit; i.e., the robot moves and guides the direction as long as the partner complies within pre-established force limits. The force gain $K_f$ can be used to adjust the limiting guidance force $F_d$. This guidance limit force is based on a previous study in which we determined the minimal interaction force for guiding the partner's motion in this framework [16].

### C. Combining Physical and Cognitive Interaction

Mutual understanding of the human–robot couple is important for establishing a successful interaction. On one hand, the human expects the robot to be a guide, a teacher who will provide support and encouragement during training. On the other hand, the robot perceives motions and forces (as described in the previous section), which alone are insufficient to fulfill student expectations. Thus, we propose to base the robot guidance on the learning process via an adaptive controller that utilizes the interaction data for modifying the robot behavior over a long-term interaction. We focus on the dance skill training in the first stage of motor learning, *cognitive learning*, where an understanding of the skill is developed through practice of motions, *"the learner performs successive approximations of the task"* [14]. Furthermore, considering an error-based internal model formation of a motor skill proposed in [12], we can assume that the partner would correct future motions based on feedback given by the robot in both physical and cognitive manners.

During this cognitive learning stage, the human partner requires feedback and appropriate assessment on a cognitive level, which are shown to be important factors for significant learning [3]. Based on this idea, we propose a dynamic feedback of performance, attaining to the partner with a KP that would help to adjust motions. Moreover, we developed an assessment method that adjusts the difficulty based on the practice count.

The performance measurement can be used not only to influence the cognitive level, but also to adapt the robot dynamic response. The principle behind this assessment method is supported by studies in haptics, where functional task difficulty is shown as one of the keys for enhancing haptic training [22]. Therefore, we developed enhanced dynamics adjusted by the partner's ability, unlike the previous studies of dynamic adaptation, where the enhanced dynamics of the task is removed after training to test the real task creating a temporal *after effect* (see [23]). Coupled dancing entails a continuous dynamic change in the couple, which would always remain the main task. In other words, the dynamics of the task itself vary as the human adapt to it, therefore the training performance should remain for the real task.

A high-level control is presented in Fig. 3, featuring the long-term interaction in the PT algorithm. Which simultaneously varies the control parameters of the robot and gives cognitive feedback through an interactive face that communicates the KP and KS to the human. The low-level control is as described in the previous section (II-B), with a motion generator (MG) that adapts trajectories to the students, defined as

$$\dot{X}_d = \left[ \mu_s \left( k_s \dot{X}_d^* \right) \right], \quad (4)$$

for an original desired trajectory from database $\dot{X}_d^*$, where $k_s \in \mathbb{R}^{n_{dof}}$ represents a user-based adjustment (set based on step lengths and height). And kinematic and mechanical limits together with an emergency stop command is issued via a discrete stop signal $\mu_s \in \mathbb{R}^{n_{dof}}$.

### III. PROGRESSIVE TEACHING

The control scheme here presented creates an adaptive control that translates the robot overall response from a highly damped behavior with higher interaction forces for novice partners (within the limits of the maximal force) to a less damped response with lower interaction forces for advanced partners. This type of response mimics the response that occurs in a human–human interaction (HHI) during social dancing, where the leading force is smaller for an experienced partner who recognize faster motions and follows the rhythm, according to professional dancers. Consequently, the robot behavior was designed for challenging users to improve from any initial state toward a motion that is more precise, volitional, and less constrained by the robot. This was devised through a PT methodology for assessing the development of the user over a long-term learning process. The methodology based on Piagets theory of cognitive development which states that knowledge is constructed based on experiences related to mental, biological and physical stage of the development [24]. Concepts of Piagets theory are currently applied in the assessment of skills in dance and sports, where levels or rankings are used to set goals, devise assessing methods, and classify the learners into groups according to their experience or age [25].

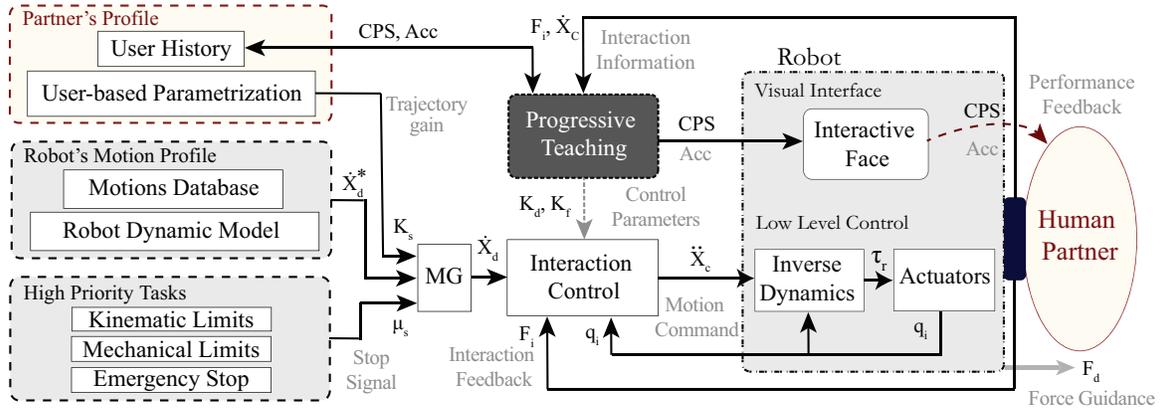

Fig. 3. High-level control of the Dance Teaching Robot, featuring interaction control with adaptive control parameters driven by the PT algorithm.

### A. Progressive Scoring for Learning Assessment

A performance evaluation system was developed based on information from the trajectory error obtained by comparing the actual path executed by the couple with the desired trajectory obtained from a set of motion–captured data of professional dancers [16]. As the interaction control was designed for guiding through desired velocities, the assessment was performed in the velocity space. We introduce a system of score zones, in which velocity error samples obtained from the interaction during practice are classified into a colored scale according to the deviation from the ideal velocity. The score zones for an example of error sampling are shown in Fig. 4(a), where the scale ranges from blue (lowest error) to grey (highest error). The velocity error is calculated as

$$e_n(k\tau) = \sqrt{\left(\dot{X}_{d_n}(k\tau) - \dot{X}_n(k\tau)\right)^2}, \quad (5)$$

$$E(k\tau) = \sum_{n=1}^{n_{dof}} e_n(k) W_n, \quad (6)$$

for an $e_n(k\tau)$ error of the $n^{th}$ dimension of the robot motion space, where $n = 1...n_{dof}$. $\dot{X}_{d_n}$ denotes the desired motion velocity and $\dot{X}_n$ the measured velocity at instant of time $k$ for $k \geq 0$ with a sampling period $\tau$. The general error $E(k\tau)$ is

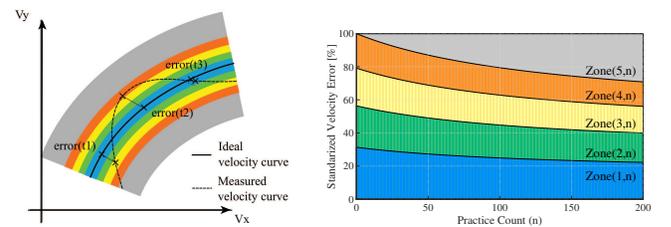

(a) Velocity error sampling under the score zone scheme.

(b) Example of the scoring zones adjusting over time (practice count).

Fig. 4. Progressive teaching basic tightening strategy for adjusting the difficulty level in the task.

the result of the weighted sum of errors $e_n$ of each dimension with weights $0 \leq W_n \leq 1$ constrained to $\sum_{n=1}^{n_{dof}} W_n = 1$.

These score zones are adjusted over time based on the student's progress, getting narrower as practice count for dance figures increases. It presents a continuous method of classification of students based on their experience. The score zone limits are calculated as

$$Zone(x, n) = c_3 \left(\frac{1}{c_2 n + 1} + 1\right)\left(\frac{1}{x + c_1}\right), \quad (7)$$

where $x$ represents an integer in the interval $1 \leq x \leq 4$ corresponding to the score zones blue through orange, and $n$ denotes the number of performed practices of the type of figure in evaluation. The variables $c_1$, $c_2$, and $c_3$ are curve adjusting parameters. The final $Zone_5$ area colored grey encompasses all errors outside the limit of $Zone_4$.

Although several methods could be used to vary the scoring system, this equation was designed aiming on convenience and simplicity. The color zone limits were selected to change hyperbolically, such that for smaller values of $x$ (closer to the ideal trajectory) the zones would be wider. The parameter $c_1$ controls the relative width between each of the colored zones. Thus, the performance evaluation forgiveness can be adjusted by changing the width ratio of zones close to the ideal trajectory (blue, green) to zones that are far from the ideal trajectory (orange, grey). The parameter $c_2$ controls the overall scale of the zones, and $c_3$ adjusts how sensitive the score zone widths are to the practice count $n$, i.e., for a higher number of practices, the trajectory error must be lower to achieve a high score. Finally, the term relating to the number of practices was also designed as a hyperbola, so that as $n$ goes to infinity the term approaches to 1 and the score zone widths converge to finite values. Fig. 4(b) shows how the practice count of a figure affects the score zones width. The desired behavior of the robot is reflected; the incentive for improving in the task is seen as a stricter assessment for increase in the practice count.

### B. Dynamic Performance Feedback

Considering progressive scoring by error zones, we introduce a two-part performance feedback for students using three measurements: on-line performance, cumulative performance score (CPS), and a final score. The first two give the students dynamic KP. Therefore, they are displayed to the students in a simple and intuitive way, i.e., based on a colored scale, whose settings are shown in Table I. The third measurement gives only knowledge of results (KR) at the end of the practice sessions.

The first measurement, on–line performance, considers motion errors (6) for one performed figure and classify them into a color zone. Subsequently, the corresponding color is displayed to the student as a highlighted bar on the interactive screen. The CPS is a general assessment that considers the progress over time, i.e., the student's CPS is recorded from the first interaction and changes with further practice. This is calculated as

$$cps(n_s) = \sum_{k=1}^{n_s} \alpha_z Z(k\tau) \quad (8)$$

$$Z(k\tau) = f(E(k\tau)), \quad (9)$$

where $Z(k\tau) \in \mathbb{R}$ is a real number defined by a $x$ number of zones part function of the general error $f(E(k\tau))$. With positive values in zones close to the desired trajectory and negative values for zones that are away from the desired trajectory (Table I defines the used function). $\alpha_z \in \mathbb{R} > 0$ represents a positive defined velocity factor for adjusting how the CPS rate changes. The score zone points $Z(k\tau)$ are accumulated from the first sample $k = 1$ to the current count of samples $n_s$. Thus, it provides a measurement of the student's performance throughout the whole training. The final score $CPS$ is defined as,

$$CPS(n_s) = \begin{cases} cps_m & : cps(n_s) \geq cps_m \\ cps(n_s) & : -cps_m < cps(n_s) < cps_m \\ -cps_m & : cps(n_s) \leq -cps_m \end{cases} \quad (10)$$

where $cps_m$ limits the function range to a specified value. This result is presented by changing the robot's face color for easy understanding of the students current progress (see Fig. 6). The third measurement, final accuracy is a general score of the overall practice calculated as a weighted average of the performance in each type of dance figure

$$Acc = (\sum_{x=1}^{h} \mu_x n_x)/(n_T \mu_h), \quad (11)$$

where $n_x$ represents the number of samples registered in each of the score zone $x \in \mathbb{R}^5$, $\mu_x$ a weight corresponding to each zone. $n_T$ represent total samples in the practice, and $\mu_h$ the highest weight, for $h$ number of zones.

Frequency of feedback to the partner is defined for the higher level of control (PT upper loop and CPS feedback in Fig. 3), through the time execution of a discrete task (a dance figure), thus the PT period is defined by $T_{PT} = T_{figure}$, which depends of the rhythm. On the other hand, the low–level interaction control loop has a period $T_c = 1\mu s$ for actuator and sensor feedback.

### C. Progressive Teaching Adaptation and Response

PT is achieved through the progressive score system combined in the CPS, which assess the general performance of each partner on the long-term pHRI. We developed an adaptive control considering the following desired adaptation of the robot: the robot should allow experienced partners to move by themselves, but provide stronger guidance to novice partners. This is presented as follows:

$$\Gamma^\star(n) = \left(\frac{1}{cps_m n_s}\right) \sum_{n=1}^{n_s} CPS(n), \quad (12)$$

$$\Gamma(n) = \left(\frac{\Gamma^\star(n) - \Gamma^\star_{min}}{\Gamma^\star_{max} - \Gamma^\star_{min}}\right), -1 \leq \Gamma(n)^\star \leq 1 \quad (13)$$

$$K_d = k_{d_{min}} + (k_{d_{max}} - k_{d_{min}})(1 - (\Gamma(n)/\alpha_d)) \quad (14)$$

$$K_f = (1 - (\Gamma(n)/\alpha_f)), \quad (15)$$

where the learning gain $\Gamma^\star(n)$ denotes the area covered by the CPS, as a stable measurement of learning response over time, standardized to the maximum $cps_m$ for the number of samples $n_s$. $\Gamma(n)$ represents the learning gain factor for adapting the robot responses over a range $0 \leq \Gamma(n) \leq 1$. The damping gain $K_d$, ranges between established limits $k_{d_{min}} \leq K_d \leq k_{d_{max}}$ as in (14), where the factor $\alpha_d$ allows reduction of the sensitivity of the response to performance changes, i.e., the robot damping will decrease $\alpha_d$ times slower than the partner's performance improvement.

The interaction gain corresponding to force $K_f$ varies in the range $0 \leq K_f \leq 1$, i.e., the desired interaction force is adapted from an initial value, reducing it if there is an improvement in the interaction, with the goal of achieving a desired interaction force of 0, as shown in (15). The sensitivity factor $\alpha_f$ adjusts the rate of change, as in the damping case. The force adaptation sensitivity is set to be much smaller than the damping one, as changes in force interaction affect the overall performance significantly more, i.e., only experienced and skilled dancers would perform with a small contact force as motion timing would be well synchronized.

## IV. EXPERIMENTAL SETUP FOR THE DANCE TEACHING ROBOT

### A. Robot Parameters

For practices with novice students, we focused on lower body motions, setting equal error gains for the lower body and 0 to the upper body in (6), thus, $W_n = [1/3, 1/3, 1/3, 0, 0, 0, 0]^T$ for $\dot{X}_n = [\dot{X}_x, \dot{X}_y, \dot{X}_\phi, \dot{X}_{q1}, \dot{X}_{q2}, \dot{X}_{q3}, \dot{X}_{q4}]^T$. Table I defines the $Z(k\tau)$ values for each color zone, CPS thresholds for dynamics feedback using the interactive face, and $\mu_x$ values for accuracy results. The parameters of (7) are set to $c_1 = 7$, $c_2 = 0.07$, and $c_3 = 14$. CPS parameters are set to $\alpha_z = 0.4$ and $cps_m = 50$. Adaptive control parameters are set to $\alpha_d = 4$, $\alpha_f = 50$, for $x$ and $y$ motions, $k_{d_{min}} = 80$, $k_{d_{max}} = 130$, $F_{dx} = [-60N, 32N]$, $F_{dy} = [-34N, 34N]$; for $\phi$ (rotational motion) $k_{d_{min}} = 50$, $k_{d_{max}} = 100$, $F_{d\phi} = [-10Nm, 10Nm]$.

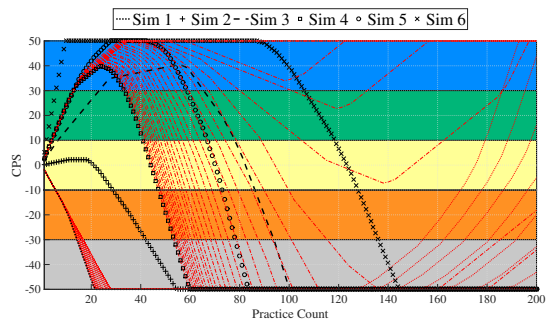

Fig. 5. Simulation of CPS behavior for different learning rates under a model of simple error-basd learning law for internal model formation. Black lines denote zero learning gains and red lines increasing learning gains.

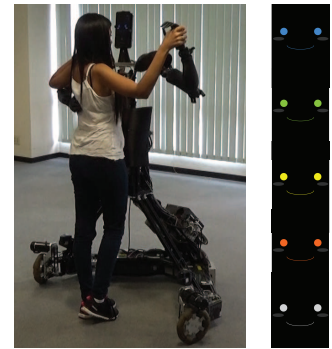

Fig. 6. Dance Teaching Robot in close contact with a human partner, and the scale of face colors for dynamic feedback of performance.

A simulation of the CPS system is presented in Fig. 5, which shows examples of the velocity error over $n = 200$ practices. In this figure, each black line denotes a different error $E(k\tau)$ used to calculate the CPS (From simulation 1 with the highest error to simulation 6 with the smallest). The learning rate simulation is performed based on the simple learning law for internal model formation proposed for the nervous system in [12], a simple gradient correction of previous error to generate future motions. The learning gain denoted by black lines, would be the extreme behavior of an internal model with learning gain 0. In each example the incentive behavior of PT is observable, initially rising toward positive score zones (green or blue), subsequently stabilizing in a zone for a certain time or saturating at the upper limit. Nevertheless, as time (practice count) advances, the tightening of score zones is reflected by the decrease in the CPS. In such circumstances the system is virtually increasing the task difficulty, as the same score achieved previously is insufficient to hold a good score. Thus, the CPS challenges the students with an enhanced task. The red line represents the expected response of students that improves over time, i.e., with learning gains > 0, which shows correction of the error over time. As practices advances and learning occurs (in reaction error improvement) the CPS rises again to positive zones. With this functional task difficulty in-troduced in the scoring we provide incentives for improvement to students.

### B. Experimental Protocol

Testing of the designed control architecture was performed with a group of 12 volunteers (6 women, 6 men), all novice in the Waltz dance, with ages in the range $20-30$. The following experimental procedure was designed for establishing a trustful relationship between students and the robot:

1) The students observe an experienced dancer performing the dance figures: the robot lacks feet, therefore, an example of the feet motions is preferable to start the learning process.
2) The students observe the robot performing the dance figures alone, for understanding of the robot motion in relation to the dance steps.
3) The students interact with the robot for the first time, adapting the robot height and arms position.
4) Students are asked to stop moving while practicing a short sequence of figures. Testing the limits of interaction forces. So that the students gain trust in the robot as they become aware that the robot motion depends of their response.
5) The students practice with the robot in a session of 20 to 30 minutes. The students do not know the dance sequence beforehand; thus, they can observe the relationship of force guidance and motion learning during the practice session.
6) The students fill a questionnaire to evaluate the human perception of the interaction.

The practice sessions continue as in step 5. Fig. 6 presents a human–DTR couple during practices.

A basic Waltz dance figure was selected, which can be performed in six different directions. Walking forward and backward and four possible combinations of *Close Changes* (CC) motions: left-forward (CCLF), left-backward (CCLB), right-forward (CCRF), and right-backward (CCRB) [10].

To investigate the PT effect on the subjective impressions of the trainees, we evaluated the psychological safety parameters of *Comfort, Peace of mind, Performance, Controllability, and Robot-likeliness* [26], as well as, a parameter of the perceived *Personal Growth* because such psychological impressions are crucial to sustain the longitudinal practice of dance for the students. Subjects were divided into two groups. One group participated in a baseline experiment, where the robot response was constant over the whole interaction (non-PT adaptations), at a predefined interaction parameter for each student set during stage 3. The second group performed with the robot executing the PT control architecture as described in section II-C.

TABLE I
COLOR ZONE SETTINGS FOR DYNAMIC FEEDBACK.

| Color Zone Scale | $f(E(k\tau))$ | CPS Color threshold | $\mu(x)$ |
|---|---|---|---|
| Blue   | +1.5 | $30 < CPS \leq 50$ | 3 |
| Green  | 0    | $10 < CPS < 30$    | 2 |
| Yellow | -1   | $-10 < CPS < 10$   | 1 |
| Orange | -2   | $-30 < CPS < -10$  | 0.5 |
| Grey   | -2.5 | $-50 \leq CPS < -30$ | 0 |

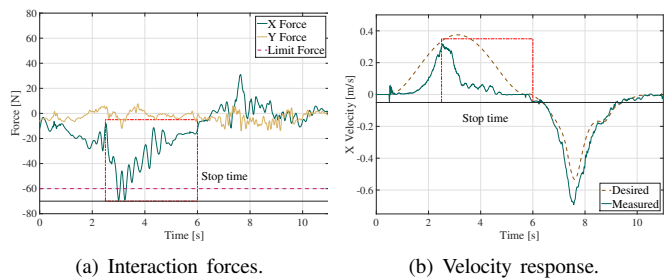

(a) Interaction forces. (b) Velocity response.

Fig. 7. Force limit interaction test example. The red areas mark the partner's stopping time, where $F_{iX}$ increases and the robot responds with a proportional decrease of the velocity. After $t = 6s$ the motion continues.

Finally, the interaction is synchronized through a rhythm given by voice-over commands that tell the partner the counting for each step. Moreover, the robot *Rise & Fall* motion is synchronized with the directional change over the planar motions, which improves motion timing of the dance, as we have shown in [27].

## V. USER-STUDY RESULTS

### A. Synchronized Motion

The initial interaction test, described in stage 3 of the experimental procedure, allows for the analysis of the robot interaction control during the close contact motion. Recordings of this test are given in the supplemental material (first part of the video). Fig. 7 presents an example data where the subject was asked to stop moving at $t = 2s$. With this procedure, the interaction force in the $X$ direction increases (see Fig. 7(a)), resulting in the robot velocity deviating from the desired trajectory as the force approaches the limit (see Fig. 7(b)). Finally, the robot stops until the subject recommences the dance. This test shows that synchronous motion with a partner is possible using the designed interaction controller limited by the desired interaction force, where the robot intends to guide as long as the partner complies with it.

### B. Progressive Teaching and Subjective Perceptions

The skill learning process is unique to each person and varies over time. Such differences are visible in the subjects' performance results shown in Fig. 8.

In the PT group results (Fig. 8(a)) some subjects, e.g., student 3, adapt easily from the beginning and improve quickly, while others, e.g., student 6, initially show poor performance, and require a longer practice period to improve. Nonetheless, the PT adaptation allows the robot to guide both types of students during the initial stage of the skill learning process. In these results 5 out of the 6 subjects ended with mid-high positive CPS, a result that reflects an improvement in skill. Furthermore, CPS fluctuations resulting from the enhanced dynamics produced by the variance in task difficulty are visible among all subjects; displaying the desired challenges to the student of the PT behavior.

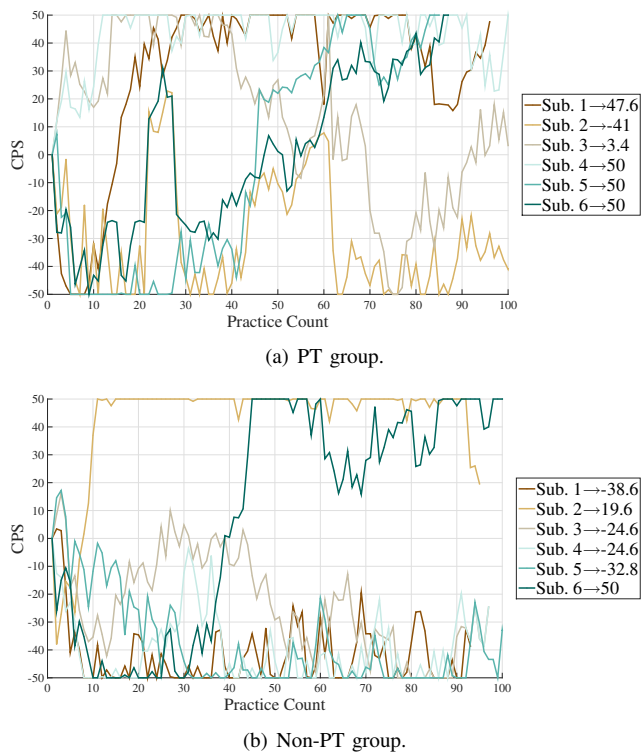

(a) PT group.

(b) Non-PT group.

Fig. 8. CPS variations for 12 subjects divided in two groups. The proposed PT adaptive control in (a), and a constant interaction control in (b), both showing final CPS result in the right-side legends.

On the other hand, non-PT group results (Fig. 8(b)) also display differences in student performances. However, in contrast with the PT group, 4 out of the 6 subjects ended with low negative scores. Furthermore, CPS remains either in positive or negative areas with less fluctuation than the PT group; leading to the conclusion that PT interactions increases the challenge in the task and rate of learning.

Concerning the impressions of the subjects, the PT training showed higher levels of *comfort*, *performance*, and *peace of mind*, as shown in Fig. 9. From the comfort and performance levels, the PT adaptive algorithm is perceived to provide a more comfortable interaction with better perceived performance of the robot. This could be due to force guidance and damping parameters that are better matched with the unique ability displayed by each partner. The peace of mind levels show a better perceived safety in the PT group. This is a result of the robot adjusting its behavior toward each partner, which may be easing the mental load of the student. This is in contrast to the baseline experiment where a constant robot behavior forces the student to totally adapt to the robot. Other measurements of robot-likeness (Non-PT = 3.26 and PT= 3.11), controllability (PT= 6.61 and Non-PT= 6.17), and personal growth (Non-PT = 5.87 and PT= 5.56), did not show significant differences.

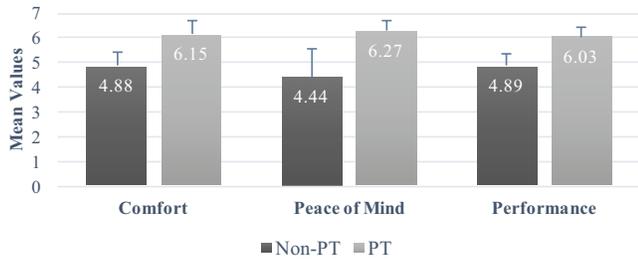

Fig. 9. Comparision of PT and Non-PT groups by one-way ANOVA. Significant difference at the $p < .01$ level for the two conditions was found for Comfort ($F(1,9) = 15.69, p < .01$), Peace of mind ($F(1,9) = 11.81, p < .01$), and Performance ($F(1,9) = 19.91, p < .01$).

## VI. Discussion

We introduced an interaction methodology for robot teaching during close contact pHRI by combining cognitive and physical interactions through a dynamic assessment of performance and an adaptive force guidance system. As an example of the methodology, robot guided interactions while teaching dance figures was demonstrated. Moreover, through the proposed *progressive teaching* (PT) algorithm the robot's behavior reflected an understanding of the human's ability during the initial stage of skill learning. Better performance and perception from students was observed compared with a non-adaptive constant behavior. Through the PT method, student encouragement for improvement was achieved by adjusting the physical impedance of the robot, resulting in fluctuation in the CPS which was overcome by the subjects. This method helps to understand the human partner state during the learning process, thus improving the overall HRI. It could therefore, be applied in other teaching scenarios such as sports and rehabilitation.

Nevertheless, the adaptive behavior of the robot can be further improved by considering time gaps between practices as a forgiveness factor in the PT algorithm. Furthermore, the current developed system relies on a long-term interaction with the robot to adjust its behavior. A fast adaptation method to the unique compliance characteristics of each student remains a future goal.

## Acknowledgment

We thank to the subjects for spending time training with the robot and the dance teachers for their support.